\documentclass{cai}
\usepackage{graphicx}
\usepackage{algorithm}
\usepackage{amsmath}
\usepackage{algpseudocode}
\usepackage{amssymb}
\usepackage{amsfonts}
\usepackage{mathrsfs}

\newtheorem{Theorem}{Theorem}

\newtheorem{Definition}[Theorem]{Definition}

\newtheorem{Proposition}[Theorem]{Proposition}

\newtheorem{Corollary}[Theorem]{Corollary}

\begin{document}
\label{firstpage}

\title{A resource - efficient model for  Deep Kernel Learning}

\author[Luisa D'Amore]
       {Luisa D'Amore}

\affiliation{Department of Mathematics and Applications\\
University of Naples Federico II\\
Complesso Monte Sant'Angelo\\
80126 Napoli}

\email{luisa.damore@unina.it}

%
%

\noreceived{} \nocommunicated{}

\maketitle

\begin{abstract}
According to the Hughes phenomenon, the major challenges encountered in computations with learning models  comes from the scale  of complexity, e.g. the so-called curse of dimensionality. 
There are various approaches for accelerate  learning computations with minimal loss of accuracy. These approaches range from model-level to implementation-level approaches. To the best of our knowledge,
the first  one  is rarely used in its basic form. Perhaps, this is due to  theoretical understanding of mathematical insights of model decomposition approaches, and thus the ability of developing mathematical improvements has lagged behind. We describe a model-level decomposition approach that combines both the decomposition of the operators and the decomposition of the network. We perform a feasibility analysis on the resulting algorithm, both in terms of its accuracy and scalability. 
\end{abstract}


\begin{mathclass}
AB-XYZ
\end{mathclass}

\section{Introduction}

Predictive Data Science is the paradigm shift of computational science tightly integrating numerical simulations with algorithms and techniques having the capability of extracting insights or useful knowledge from data (also known as learning-from-examples).  Predictive Data  Science - revolutionizing decision-making for high-consequence applications in science, engineering and medicine -  aims  not only to  reproduce with high-fidelity the real world, but also to predict its behaviour  in situations for which the numerical simulation has not been specifically validated.  Machine learning (ML) is part of predictive data science, dealing with statistics, algorithms, and scientific methods used to extract knowledge from data \cite{Poggio2020,Poggio2003}. Various types of ML algorithms such as supervised, unsupervised, semi-supervised, and reinforcement learning exist in the area. In addition, deep learning (DL), which is part of a broader family of machine learning methods, can analyze the data on a large scale by learning from subtasks. DL is having an immense success in the recent past leading to state-of-art results in various domains (biotechnology, finance, ....). Anyway, the same flexibility that makes DL excellent at modeling diverse phenomena and outperforming other models (which indeed depends on large amount of data and computation) also makes it dramatically more computationally expensive. If the current trend continues, the growing computational cost of DL will become technical and economically prohibitive \cite{Thompson}.   The continuing appearance of research on distributed learning is due to the progress made by specialized high-performance architectures.  But the computational needs of DL scales so rapidly that it will quickly becomes computationally constrained again.\\ 
We present a model-level technique for tackling intensive DL problems which relies on the ideas of Kernel decomposition approach. We call it D$^3$L.  D$^3$L involves  data  reduction,  localization of the predictive function and reduction of the  error function. Main feature of D$^3$L is that local  error functions are  suitably modified, by imposing a regularization constraint in order to enforce the matching of their solutions  between adjacent subproblems. As a consequence, instead of solving one DL problem, we can solve several smaller problems improving the accuracy-per-parameter metric. Most importantly,  subproblems can be solved in parallel,   leading to a  scalable algorithm where the workers locally exchange parameter updates via a nearest-neighbourhood communication scheme, which does not require a fully connected netwok.  
\subsection{Related Works}

\noindent  Resource-efficient DL research
has vividly been carried out independently in various research communities including the machine
learning, computer arithmetic, and computing system communities. There are various approaches to compress or accelerate DL methods with minimal loss of precision \cite{BEN-NUN,Hestness,Mayer,Mayer2017,Shazeer,yucheng}. Among them, the model - level techniques aims at  reducing the problem size to fit the DL models to resource-constrained systems. On the contrary, implementation-level techniques aims at improving the computational speed of DL operations \cite{junkyulee}. These approaches range from the fine to the coarse grained parallelism.  The first one corresponds to the standard fine-grained concurrency of the floating point operations (it exploits concurrency inside the parts that represent the main computational bottlenecks of the neural networks layers to enhance the overall performance of the whole algorithm). In the realm of DL this approach is often referred to as \emph{concurrency in operators}.  To implement fine-grained parallelism effectively, specialized hardware such as GPUs and TPUs are commonly used. These hardware accelerators are designed to handle a multitude of parallel operations simultaneously, making them well-suited for DL workloads. Finally, parallel execution techniques such as multiprocessing and multithreading are employed to execute parallel operations concurrently. This approach significantly reduces the training time for deep neural networks, allowing for the training of larger and more complex models. \\
 The second one is based on the problem decomposition and, in contrast to the fine-grained parallelism,  introduces concurrency at a coarser level of computation. In the realm of DL, this approach is known as\emph{concurrency in network}, and involves partitioning the computational workload in various ways: by input samples (data parallelism), by network structure (model parallelism), and by layer (pipelining)\cite{Dutta,Jager}. Data parallelism is a straightforward approach to parallelization. This method dates back to the first practical implementations of artificial neural networks \cite{Zhang}. Basically, data parallelism partitions the dataset  among processing units having a copy of the full model. Each unit calculates the gradients of different subsets, independently, and uses these gradients to update the global  model concurrently.  Most DL frameworks support data parallelism (Chainer \cite{Chainer}, PyTorch \cite{PyTorch}, Caffe2 \cite{Caffe}, CNTK \cite{CNTK}).
 In the model parallelism, data are copied to all processors,  each unit has a portion of the model, and works in cooperation with others to do the calculation for one subset. Subsequently,  different stages of the calculation of the global model  are executed in pipeline  on different processors. In  layer pipelining, different layers of the neural network are processed concurrently, with data flowing through the network in a pipeline fashion. It is a strategy to reduce the latency in DL inference. \\
 Model parallelism has lower communication requirements, but because of its pipeline data dependency, model parallelism suffers from under utilization of the computing elements, while data parallelism does not have the data dependency issues, but requires heavier communication across processing units due to the synchronization with the other processing units.  The combination of multiple parallelism schemes can overcome the drawbacks of each scheme giving rise to hybrid approaches. In this regard, most notable frameworks supporting  hybrid approaches are TensorFlow \cite{TensorFlow}, MXNet \cite{MXNet} and SINGA \cite{SINGA}.

\subsection{The present work}

\noindent   According to the recent survey in \cite{junkyulee},  we can summarize main aspects of significance and novelty of  D$^3$L as follows. 
\begin{itemize}
\item \emph{Parallelization Model}. We introduce an hybrid parallelism which starts with a decomposition of the global problem into overlapped smaller sub problems. On these sub domains we define local problems and we prove that the solution of the (global)  problem can be obtained by collecting the solutions of each local problem. The (global) problem
is decomposed into (local) sub problems in such a way.
    \item \emph{Framework architecture}. The proposed approach works without any parameter services, e.g. it is decentralized, but it does not need to exchange parameter updates by using an all reduce operation. 
The resulted algorithm consists of several copies of the original one, each one requiring approximately the
same amount of computations on each sub domain and an exchange of boundary conditions
between adjacent sub domains. 
 
\item \emph{Syncronization}. The  partitioning  of the computational domain requires only interactions among two adjacent subdomains. As data is flowing across the surfaces, the so called surface-to-volume effect is produced.
As the equations in the subdomains are solved concurrently the synchronization  of local solutions is imposed
iteratively.  Such synchronization guarantees the model convergence, although it may slow down the whole work. In such cases, indeed, a static and or a priori decomposition  could not ensure a well balanced work load, while a way to re-partition   the mesh so that the sub domains maintain a nearly equal number of observations plays an essential role in the success of any effective DD approach. A  dynamic load balancing algorithm  allows for a minimal data movement restricted to the neighboring processors. This is achieved by considering  a connected graph induced by the mesh partitioning whose vertices represent a sub domain associated with a scalar representing the number of observations on that sub domain. Once the  domain has been partitioned a load balancing schedule (scheduling step) should make the load on each sub domain equal to the average load providing the amount of load to be sent to adjacent sub domains (migrations step).  The most intensive kernel is the scheduling step which defines a schedule for computing the load imbalance  (which we quantify in terms of number of observations) among neighbouring sub domains. Such quantity is then used  to update the   shifting the adjacent boundaries of  sub domains which are finally re mapped  to achieve a balanced decomposition. We are assuming that load balancing  is restricted to the neighbouring domains so that we reduce the overhead processing time. Finally,  we  use a diffusion type scheduling algorithm minimizing the euclidean norm of data movement. The resulting constrained optimization problem leads to the solution of the related normal equations whose matrix is associated to the decomposition graph. 
\item \emph{Communication}.  The  approach we introduce is  extremely easy  to implement on emerging parallel architectures. This is due to the  ability of exploiting  multiple levels of parallelism depending  both on the granularity of the operations and on the mapping on the target architecture. In particular, theoretical performance analysis in terms of the scale up factor tell us that  if the application  needs to reach a prescribed time-to-solution (strong scaling), we can exploit the high performance of emerging GPU. In this case, the number of processing elements can increase while  local problem size  is fixed according to memory constraints of GPU;  the scale-up factor increases with  a fixed  surface-to-volume ratio. On the other hand, for computationally intensive applications, it is preferable to exploit the weak scaling of  clusters of distributed memory multiprocessors. Indeed, by fixing the number of processors  while the local problem size may  increase according to the requirements of the application, the scale-up factor is kept constant while the surface-to-volume ratio decreases supporting the overall efficiency.

\end{itemize}

\noindent It is worth noting that the efficient implementation of any  model-level
techniques on given compute platforms is essential to improve physical resource efficiency. In this regard, we are mainly interested in investigating the integration of the STRADS interface into the framework  \cite{junkyulee}. Here we start presenting the feasibility analysis of the proposed approach and validate its scalability using the high-performance hybrid computing architecture of the SCoPE (Sistema Cooperativo Per Elaborazioni scientifiche multidiscipliari) data center, located at  University of Naples Federico II. The  architecture is composed by 8 nodes consisting of distributed memory DELL M600 blades. The blades are connected by a 10 Gigabit Ethernet technology and each of them is composed of 2 Intel Xeon@2.33GHz quadcore processors sharing the same local 16 GB RAM memory for a number of 8 cores per blade and of 64 total cores. We study the performance  by using Parallel Computing Toolbox of MATLABR2014b. \\

\noindent The rest of the article is described as follows. Section \S 2 introduces ML and DL as identification problems while in Section \S 3 by restricting the analysis
on Reproducing Kernel Hilbert Spaces we cast DKL (Deep Kernel Learning) problems into a framework
exploiting the connection with the theory of inverse and ill posed problems.  Finally, we formulate DKL problems as Concatenated Tikhonov Regularization functionals ($\mathcal{CTR}$).  Section \S 4 is focused on the new approach to $\mathcal{CTR}$ reduction, while the algorithm is presented in Section \S 5 with its performance analysis. Section \S 6 discusses main outcomes of this analysis.
 
\section{Basic mathematical concepts of learning from examples}
\noindent ML can be regarded as an \emph{identification} problem. Identification problem is indeed a problem of model formulation. According to \cite{Linz} we say that a model is identified if it is in a unique form enabling unique estimates of its parameters to be subsequently made from available data. In line with this concept we review basic mathematical definitions of ML \cite{Poggio2020,Poggio2003,Rosasco2004,Rosasco}. 
\begin{Definition}[ML - Problem I]
We are given the input space $X$ which we assume to be  a compact subset of $\Re$, the output space $Y$ which is a subset of $\Re$ contained in $ [-M,M]$ for some $M \geq 0$ and  the (training) data set $S:=(x_{i},y_{i})$, for $i=1,\ldots,N$, which are samples in $X \times Y$.
Given the  data set $S$ the aim of any (supervised) learning problem is to find the function  $\Phi: X \rightarrow Y $ (also known the predictor) which is able to well estimate any new  output $y \in Y$ once a new input  $x \in X$ is given (or, function  $\Phi$  generalizes the output from unseen input). 
\end{Definition}
\noindent Solving an identification problem  means finding a way of incorporating in the model  information coming from data. We recognize that identification problem is logically prior to data estimation problem. More precisely, we could not understand from the data set which specific relationship it is representing unless we get a particular form of it. In other words, as we will see later, the typical approach to the solution of the problem is to define a metric (the loss function) depending on the data and on the model  $\Phi$, and solve an approximation problem. In order to understand how such approximation problem  comes out, we recall data estimation problems, already introduced by R. E. Kalman, in his pioneering work in 1960 \cite{Kalman}. 
\begin{Definition}[Data Estimation]
Given the points $$ (t_i, y_i), \quad i = 1,2 \quad t_i \in [0,T]$$ where
$y_i=x_1(t_i) +\epsilon_i ,$ to calculate $x_1(\widetilde{t})$, $\widetilde{t}\in [0,T]$. \noindent Depending on the position of $\widetilde{t}$ with respect to $t_1$ and $t_2$ in~\cite{Kalman} this problem was characterized as follows:
\emph{
\begin{itemize}
  \item $t_1 < \widetilde{t} < t_2$: data smoothing (fitting of data)
  \item $\widetilde{t}= t_i$: data filtering
  \item $\widetilde{t}>\{t_1,t_2\}$: data prediction (data mining)
\end{itemize}
and, in general, it was called data estimation. \\}

\end{Definition}
\noindent In numerical computing  data estimation is solved as  an approximation problem. \emph{C. F. Gauss} in 1795, at age 18, in his study of the orbits of the planets stated that\cite{Gauss}:
 \begin{quote}
 "\emph{[...] measurements are affected by errors and so are all obtained from these computations, therefore, the only way to get information about the problem at hand is to compute an approximation of the nearest and most practicable solution possible. This can be done by using a suitable combination of the experimental measurements, which must be in number than those of the unknown parameters, and starting from an approximate knowledge of the orbit (to be calculated), which will be corrected in order to describe as accurately as possible the experimental observations."}
\end{quote}
Gauss focused on  the main ingredients needed for the computation of the solution of an approximation problem through:
\begin{enumerate}
  \item the use of experimental measurements in a number higher than that of the unknown parameters;
  \item the identification  of the model linking the quantities and the known unknowns;
  \item the calculation of the minimum distance between the known values and those obtained  by solving the model.
\end{enumerate}
 Following Gauss, we need to refine ML Problem I given in Definition 2.

\section{Learning form examples: a large scale inverse ill posed problem}
\noindent  Deep Learning is the enhanced version of ML where  models deal with complex 
tasks by learning from subtasks. In particular, several nonlinear functions are stacked in hierarchical architectures to learn multiple levels of representation from  input data (the higher-level features are defined in terms of lower-level ones).  Each function transforms the representation at one level into a much more abstract representation at a 
higher level.  The core building block of DL are mathematical functions called artificial neurons \cite{LeCun}.  
\begin{Definition} [Artificial neuron]
An artificial neuron with weights $w_1,\ldots, w_N \in \Re$, bias $b \in \Re$ and (activation) function $\rho: \Re \rightarrow \Re$ is defined as the function $f: \Re^N \rightarrow \Re$ given by  
\begin{equation}
    f(x_1, \ldots x_N) = \rho \left ( \sum_{i=1}^{N}x_i w_i-b \right) 
\end{equation}
\end{Definition}
\noindent The simplest form of deep networks is the Deep Feedforward network (or deep neural network, DNN)  described as a collection of artificial neurons which are organized in layers. Neural networks are basically made up of three layers: an input layer, a hidden layer, and an output layer. Adding two or more hidden layers to a traditional neural network we obtain the DNN. Each layer has a set of units. The units between adjacent layers are inter-connected and each connection is associated 
with a weight parameter. In each layer the input vector first goes through an  affine linear transformation  and then pass through the  activation function \cite{Berner,LeCun}. 

\begin{Definition} [DL -  Problem I]
Let $d \in \mathcal{N} $ be the dimension of the input layer, $L$ the number of layers, $N_0:=d$, $N_l$, where $l=1,\ldots L$ the dimensions of the hidden and last layer, $\rho$, an activation function and, for $l=1,\ldots, L$ $T_l$ be the affine linear functions 
\begin{equation}
        T_l\mathbf{x}= W^{(l)}\mathbf{x}+b^{(l)}
\end{equation}
with $W^{(l)} \in \Re^{N_l \times N_{l-1}}$ being the weight matrices, $\mathbf{x}=(x_1, \ldots,x_d)$, and $b^{(l)}\in \Re^{N_l} $ the bias vector of the $l$-th layer. Then the composite function $\Phi: \Re^d \rightarrow \Re^{N_L}$ given by

\[ 
\Phi: = T_L \circ \rho \ldots\circ \rho \circ T_1  
\]
such that
\[ 
\Phi(\mathbf{x}) := T_L \rho(T_{L-1} \rho ( \ldots \rho (T_1(\mathbf{x}))))
\]
is a (deep) neural network of depth $L$.   The activation function is applied at each hidden unit to achieve the 
nonlinearity of neural network models. Commonly used activation functions include sigmoid, hyperbolic tangent, and rectified linear unit (ReLU) functions \cite{Berner,LeCun}. 

\end{Definition}
\subsection{The Learning inverse ill posed problem}
\noindent Kernel methods have been successfully applied to a wide variety of learning problems. These methods map data from the input space to a Reproducing Kernel  Hilbert Space(RKHS) by using a kernel function which computes the scalar  product between data points in the RKHS. By restricting the analysis on RKHS we cast  Deep Learning models into a functional analysis framework exploiting the connection with the theory of Tikhonov Regularization \cite{Rosasco2004,Scholkopf}.\\
\noindent Formally, an RKHS is a Hilbert space of functions on some domain  in which all evaluation functionals are bounded linear functionals \cite{Aronszajn}. 

\begin{Definition}[RKHS]
Let $H$ be an Hilbert space of functions from $\Omega \subset \Re^N$ to $\Re$, equipped with the scalar product $<\cdot,\cdot>$. $H$ is a RKHS if it exists a  function 
\begin{equation}
    K: \Omega \times \Omega \rightarrow \Re 
\end{equation}
which is  called a Reproducing Kernel (RK) of $H$,   satisfying:
\begin{enumerate}
    \item $\forall \, \mathbf{x} \in \Omega,\, k_x= K(\cdot ,\mathbf{x}) \in H$;
    \item $\forall \, \mathbf{x} \in  \Omega, \, \forall \, f\in H, \, <f, K(\cdot, \mathbf{x})>= f(\mathbf{x})$.
\end{enumerate}
\end{Definition}
\noindent The RK is always symmetric and positive definite. Every RKHS has a unique RK. Conversely for every positive definite function $K$ there exists a unique RKHS with K as its RK. 
 \begin{Definition}[ML in RKHS - Problem II]
Let $H$ be an Hilbert space of functions from $X$ to $\Re$. Given the setup as in Definition 1 of ML Problem I,  the (supervised) learning problem concerns the computation of the predictor $\Phi$ in an hypothesis space $H$ which is a RKHS  on the set $X$.  
\end{Definition}
\noindent Following result leads to solution of ML in RKHS - Problem II for noiseless data (e.g. data interpolation) \cite{Bohn,Cho}.
\begin{Theorem} [Representer Theorem]
The function $$\Phi = \sum_{k=1}^{N}\alpha_kK(\cdot,x_k)$$ is the unique minimizer of the Hilbert space norm in $H$  under all functions $f \in H$ such that 
$$f(x_i) = y_i$$
The coefficients $\alpha_k$ can be calculated form the linear system 
$$\mathbf{A} \alpha = \mathbf{y}$$
where $A_{ij}= K(x_i,x_j)$, $\alpha= (\alpha_1, \ldots \alpha_N)^T$,  $\mathbf{y}=(y_1, \ldots,y_N)^T$ and $\mathbf{A} \in \Re^{N \times N}$. 

\end{Theorem}
\noindent This result states that ML Problem II in RKHS leads to the inverse problem consisting in computing the coefficients $\alpha_1, \ldots\alpha_N$ inverting the kernel matrix $\mathbf{A}$;  since  $\mathbf{A}$ is positive definite,  it is also invertible, and the solution is unique.  
 \begin{Definition}[ML in RKHS - an inverse problem]
Let $H$ be an Hilbert space of functions from $X$ to $\Re$. Given the setup as in Definition 1 of ML Problem I,  the (supervised) learning problem consists in the solution of the linear system 

\begin{equation}\label{systemA}
  \mathbf{A} \alpha = \mathbf{y} 
\end{equation}
where $\mathbf{A} \in \Re^{N\times N}$ is the kernel matrix of $H$. 
\end{Definition}
 \noindent Solving inverse problems can be very challenging for the following reasons: small changes in the data values may lead to changes in $\Phi$ i.e., the kernel matrix $A$ can be very ill-conditioned (in the learning context this is the so-called overfitting problem) and the problem is ill posed. 
  The characterization of ill-posed mathematical problems, dates back to the early years of the last century (J. Hadamard, 1902) and reflects the belief of the mathematicians of that time to be able to describe in a unique and complete way each physics problem.
As a result, a problem was ill posed when, from the mathematical point of view, it presents anomalies and for this reason it could certainly not correspond to a physical event. Therefore, for some years, ill-posed problems were not taken into consideration by mathematicians. The first comprehensive treatment of ill-posed problems,  is due  in 1965, to A. N. Tikhonov, which described the concept of solution for ill-posed problem and introduced the regularization methods \cite{Tikhonov}. This solution is obtained by solving a best approximation problem minimizing the sum of two terms: the first is a combination of the residual between data and predicted outputs (the so-called data fitting term) in an appropriate norm, and the second is a regularization term that penalizes unwanted features of the solution. The inverse problem thus leads to a nonlinear variational problem in which the forward simulation model is embedded in the residual term. Hence, regularization methods are used to introduce prior knowledge.\\ 
\begin{Definition} [RML - Regularized ML Problem III]  
Given the setup of ML Problem II  as in Definition 6, the predictor $\Phi$ is defined such that 
\begin{equation}\label{L}
 \Phi^*= arg   min_{\Phi\in H} \sum^{N}_{i=1} \mathcal{L}(\Phi(x_{i}),y_{i})
\end{equation}
 where the operator $\mathcal{L}$ is the regularization  operator measuring the goodness of fit to data. 
\end{Definition}
\noindent This viewpoint does not guarantee to compute acceptable solutions, because continuous dependence of the solution on the data (guaranteed by the regularization methods) is necessary but not sufficient to get numerical stability. In 1988, James Demmel  discussed the relationship between ill posedness and conditioning of a problem, investigating the probability that a numerical analysis problem is difficult.  In the meanwhile, P.C. Hansen,  introduced the so-called discrete ill-posed problems, to emphasize the huge condition number of rank-deficient discrete problems arising from the discretization of ill posed problems. The key point is the computation of  regularization parameters  which are able to balance the accuracy of the solution and the stability of its computation \cite{Arcucci2017,Damore2014}.\\
A classical choice for  $\mathcal{L}$ is  Tikhonov Regularization (TR). Standard TR method consists in replacing the linear system in (\ref{systemA}) with  the constrained least square problem 
\begin{equation}\label{tikhreg}
{\alpha^*}= argmin_{\alpha}  \, \mathcal{L}= argmin_{{\alpha}} \|\mathbf{A}\alpha-\mathbf{y}\|_2+\lambda \|\mathcal{\mathbf{Q}}{\alpha}\|_2
\end{equation}
where $\mathcal{\mathbf{Q}}$ is referred to as the regularization matrix and the scalar $\lambda$ is known as the regularization parameter. The matrix $\mathcal{\mathbf{Q}}$ is commonly chosen to be the identity matrix; however, if the desired solution
has particular known properties, then it may be meaningful to let $\mathcal{\mathbf{Q}}$ be a scaled finite difference approximation of a differential operator or a scaled orthogonal projection.  Finally, $\| \cdot\|_2$ denotes the L$^2$-norms in $\Re^N$. For an introduction to the solution of this kind of problems we refer to \cite{Tikhonov}.
Regularization parameter $\lambda$ influences condition number of  $\mathcal{L}$. \noindent By using Theorem 1,  the  function $$\Phi = \sum_{k=1}^{N}\alpha^{\lambda}_kK(\cdot,x_k)$$ is the unique minimizer of (\ref{L}), where  coefficients $\alpha_k^{\lambda}$ can be calculated form the normal equations arising from  (\ref{tikhreg}):
$$(\mathbf{A}^T\mathbf{A} + \lambda \mathcal{\mathbf{Q}}^T)\alpha^{\lambda} =\mathbf{A}^T\mathbf{y} $$
where $A_{ij}= K(x_i,x_j)$, ${\alpha}^{\lambda}= (\alpha_1^{\lambda}, \ldots.\alpha_N^{\lambda})^T$ and $\mathbf{y}=(y_1, \ldots,y_N)^T$. \\

\noindent By proceeding in the same way, e.g. by considering DL models in RKHS,  we  can even employ TR   \cite{Unser}
\begin{Definition}[RDL - Regularized DL  Problems]
Given the DL problem setup as in Definition 4, given the $L$ nonlinear functions $g_i(\mathbf{x}):= \rho  \circ T_i (g_{i-1}(\mathbf{x})) $, where $g_0(\mathbf{x})=\mathbf{x}$, then  the Regularized DL (RDL) consists in  computing the function $g_i^*$ such that
\begin{equation}\label{minDKL}
\{g_i^*\}_{i=1,\ldots,L} = \arg\min_{g_i \in H} \mathcal{J}(g_i)
\end{equation}
with
\begin{equation}\label{minDKLsec}
 \mathcal{J}(g_i)= \sum_{i=1}^L \Theta_i\| {g_i}\|_2    +  \sum_{i=1}^L \| \mathcal{L} (g_L \circ g_{L-1}\circ \ldots g_1(\mathbf{x}) ,\mathbf{y}) \|,
\end{equation}

\noindent where  $\mathcal{L}$ is defined in (\ref{tikhreg}).
\end{Definition}
\noindent In particular, $L$-layer  Deep Kernel Learning (DKL) methods  are hybrid DL Problems on RKHS, which combine the flexibility of kernel methods with the structural properties of DL methods. DKL methods build a kernel by non linearly transforming the input vector of data before applying an outer kernel \cite{Aichison,Cho,Bohn,Dinuzzo,Montavon,Rosasco,Wang,Wilson}. Formally, we assume that each function  $g_i \in H_i$ for $i=1, \ldots L$ where $H_i$ are RKHS with associated kernel $K_i$.\\ 

\noindent In this case, 
the solution  is given by a linear combination of at most N basis functions in each layer and the following Concatenated Representer Theorem subsists


\begin{Theorem}[Concatenated Representer Theorem]
 Let $H_1, \ldots, H_L$ be RKHS of functions with domain $D_i$, and range $R_i \subseteq \Re^{d_i}$ where $R_i \subseteq D_{i-1} $ and $D_L=\Omega$ for $i=2,\ldots L$. Let $\mathcal{L}$ be the regularization functional  in (\ref{minDKLsec}). Then, a set of minimizer $f_i \in H_i$, $i=1, \ldots, L$  of
 \begin{equation}
     J(f_1, \ldots, f_L)= \sum_{i=1}^L \mathcal{L}(y_i, f_1\circ \ldots \circ f_L(x_i) )+  \sum_{i=1}^L \|f_l\|^2_2
 \end{equation}
 fulfills $f_i\in V_i\subset H_i$ with
 \[
V_i=span \{  K_i(f_{i+1} \circ \ldots f_L(x_j)), \cdot)  e_{k_i}   )\,,  for \, j= 1 \ldots N \,\, and \, k_i= 1, \ldots d_i\} 
 \]

 \end{Theorem}
\noindent  This result enables us to consider the Regularized DKL (RDKL) Problem instead of DKL, and,  by using TR  in (\ref{minDKLsec}), we arrive at the computation of  the minimizer of the following functional which we call  Concatenated Tikhonov Regularization ($\mathcal{CTR}$):

\begin{equation}\label{tikhreg2}
\mathcal{CTR}(f_1 \circ f_2 \circ \ldots \circ f_L)(\mathbf{x})) = \mathcal{TR}(f_1)+ \mathcal{TR}(f_2) + \ldots \mathcal{TR}(f_{L-1}) 
\end{equation}

\noindent In particular, each quadratic term in (\ref{tikhreg}) can be expressed in terms of the kernel matrix $A^l$ of each RKHS, for $l=1,\ldots, L$ 
\[
A^l_{ij}=K(f_1 \circ f_2 \circ \ldots \circ f_L(x_i), f_1 \circ f_2 \circ \ldots \circ f_L(k_j)
\]
leading to a nonlinear least square problem \cite{Dennis96} whose dimension is 
$\sum_{i=1}^L N \cdot d_i$
Interested readers can found the detailed analysis for a  two-layer least square problem  in \cite{Zhuang}.
\section{Problem Reduction}
\noindent This nonlinear least-squares problem  is typically considered data intensive with $N$ larger than $10^{10}$ and $L$ larger than $1000$ (ResNet comprises $1202$ layers and the number of layers grew about $2.3X$ each year). So how to speed the time-to-solution is an interesting and active research direction. In this context, we provide a mathematical approach based on  domain decomposition of $\mathcal{CTR}$  which starts from data decomposition then  uses a partitioning of the  solution and of the  modified  functional.

  
\noindent Starting from the  $\mathcal{CTR}$  loss functional, we define local  $\mathcal{CTR}$  functional on sub sets of data  and we prove that the minimum of the "global" functional can be  obtained by collecting the minimum of each "local" functional.  We prove that  the  "local"  inverse problems  are  equivalent. As a result, we may say that the proposed approach is loseless keeping the reliability of the global solution \cite{DD-DA,JPP}. 
\subsection{Basic Concepts}
\noindent In this section we  introduce some  concepts and notations we need to use.
We  give a precise mathematical setting for space and function decomposition then we state some notations used later. In particular, we first introduce the function and domain decomposition, then by using  restriction and  extension operators, we associate to the domain decomposition  a functional decomposition. So, we may prove the following result: the minimum of the global  functional,  defined on the entire domain,  can be regarded as a piecewise function obtained by collecting the minimum of each  local functional.
\begin{Definition}[Data Decomposition]
Let $\Omega$ be a finite numerable set such that $card(\Omega)=N$. Let
\begin{equation}\label{decom}
\Omega=\bigcup_{i=1}^{p} \Omega_i \quad card(\Omega_i)=r_i,
\end{equation}
\noindent be a decomposition of the domain $\Omega$ into a sequence of overlapping sub-domains $\Omega_i$,   where $r_i\leq N$ and $\Omega_i \cap \Omega_j =\Omega_{ij}\neq \emptyset$  when the subdomains are adjacent. 
\end{Definition}
\noindent Associate to the decomposition (\ref{decom}), we give the following:
\begin{Definition}[The Restriction and the Extension Operator]
If $\mathbf{w}=(w_i)_{i\in \Omega}\in \Re^{N}$ then
$$ RO_{i}(\mathbf{w})  := (w_i)_{i\in \Omega_i}\in \Re^{r_i}$$ is the restriction operator acting on $\mathbf{w}
$. In the same way, if $\mathbf{z}=(z_i)_{i \in \Omega_i}$, then it is
$$ EO_{i}(\mathbf{z}):= (\tilde{z}_k)_{k\in \Omega} \in \Re^N$$
where
\begin{equation}\label{exten}
\tilde{z}_k:=
\left \{
\begin{array}{cc}
  z_k & k \in \Omega_i \\
  0 & elsewhere
\end{array}
\right .
\end{equation}
is the extension operator acting on $\mathbf{z}$.

\end{Definition}
\noindent We shall use the notations $RO_{i}(\mathbf{w})\equiv \mathbf{w^{RO_{i}}}$ and
$ EO_i(\mathbf{z}) \equiv \mathbf{z^{EO_{i}}}$.\\

\noindent \textbf{Remark}: For any vector $\mathbf{w}\in \Re^{N}$,  associated to the domain decomposition (\ref{decom}), it results that
\begin{equation}\label{rest}
    \mathbf{w}= \sum_{i=1,p}EO_i \left [\mathbf{w}^{RO_i} \right ].
\end{equation}
\noindent The  summation
\begin{equation}\label{exte}
    \mathbf{w} :=\sum_{i=1,p}\mathbf{w}^{EO_i}
\end{equation}
is such that, for any $j\in \Omega$:
$$ RO_j[\mathbf{w}]=RO_j \left [\sum_{i=1,p}\mathbf{w}^{EO_i} \right ]= \mathbf{w}^{RO_j}$$

\begin{Definition}[The Functional Restriction Operator]
Let
$$ J(\mathbf{w}): \mathbf{w}\in \Re^N  \mapsto J(\mathbf{w}) \in \Re$$
be the least square operator as defined in (\ref{tikhreg}) 
\begin{equation}\label{func}
   J(\mathbf{w})= \|\mathbf{A}\mathbf{w}-\mathbf{y}\|_2+ \lambda \|\mathbf{w}\|_2
\end{equation}
defined in $\Re^N$, where $\lambda>0$ is the regularization parameter. For simplicity of notations we let $\mathbf{Q}= \mathbf{I}$, where $\mathbf{I}$ is the identity matrix. We  generalize the definition of the restriction operator $RO_i$  acting on $J$, as follows:
\begin{equation}\label{restrfun}
RO_i[ J]: J(\mathbf{w}) \mapsto RO_i[ J(\mathbf{w})]
\end{equation}
where $RO_i[ J(\mathbf{w}) ]=\|\mathbf{A}^{RO_i}\mathbf{w}^{RO_i}-\mathbf{y}^{RO_i}\|_2+ \lambda \|\mathbf{w}^{RO_i}\|_2$ and 
$$J(\mathbf{w}^{RO_i}) \mapsto  \left \{ \begin{array}{cc}
    J(\mathbf{w}^{RO_i}),& \forall \, j, \, \Omega_i\cap \Omega_i =0,  \\
    \frac{1}{2}J(\mathbf{w}^{RO_i}), & \exists \, j\,: i\in \Omega_i \cap \Omega_j
    \end{array}
 \right .
 $$
 
\end{Definition}
\noindent For simplicity of notations, and also for underlining that the restriction operator is associated to the domain decomposition (\ref{decom}) we pose:
$$ RO_i[ J] \equiv J_{\Omega_i}\quad .$$

\begin{Definition}[The Functional Extension Operator]
We generalize the definition of the extension operator $EO_i$ acting on $J_{\Omega_i}$ as
$$ EO_i: J_{\Omega_i} \mapsto J^{EO_i}_{\Omega_i}\,\,,$$
where
\begin{equation}\label{ext}
  EO_i[J_{\Omega_i}]: \mathbf{w} \mapsto  \left \{
\begin{array}{cc}
  J(EO_i(\mathbf{w}^{RO_i})) & i \in \Omega_i \\
  0 & elsewhere
\end{array}
\right . \quad .
\end{equation}

\end{Definition}
\noindent We note that the (\ref{ext}) can be  written as
\begin{equation}\label{exten2}
  J^{EO_i}_{\Omega_i}(\mathbf{w})= EO_i[J_{\Omega_i}](\mathbf{w})= J(EO_i(RO_i[\mathbf{w}]))
\end{equation}

\begin{Proposition}[Functional Reduction]
\noindent Let $\Omega=\bigcup_{i=1}^{p} \Omega_i $ be the decomposition  defined in (\ref{decom}) and let $J$ be the functional defined in (\ref{func}). It holds:
\begin{equation}\label{DD-J}
 J\equiv  \sum_{i=1,p}J^{EO_i}_{\Omega_i}\quad ,
\end{equation}
where $$J_{\Omega_{i}}: \Re^{r_i}\mapsto \Re \quad .$$
\begin{proof} From the (\ref{exten2}) it follows that, if $\mathbf{w}\in \Re^N$, then
\begin{equation}\label{exten3}
     \sum_{i=1,p}J^{EO_i}_{\Omega_i}(\mathbf{w})= \sum_{i=1,p}EO_i[J_{\Omega_i}](\mathbf{w})= \sum_{i=1,p} J_{\Omega_i}(RO_i[\mathbf{w}]) \,\,.
\end{equation}
From  (\ref{rest}), (\ref{ext}) and (\ref{exten3}) it results that
\begin{equation}\label{rest2}
      \sum_{i=1,p} \left [J_{\Omega_i}(RO_i[\mathbf{w}])\right ]^{EO_i}=\sum_{i=1,p} J \left ( EO_i( (\mathbf{w}^{RO_i})) \right ) =  J\left[\sum_{i=1,p} (\mathbf{w}^{RO_i})^{EO_i}\right ]= J(\mathbf{w}) \,\,.
\end{equation}
From  (\ref{exten3}) and (\ref{rest2}),  the (\ref{DD-J}) follows.
\end{proof}
\end{Proposition}

\subsection{TR Reduction}
\noindent Let:
\begin{equation}\label{min3glob}
\mathbf{w}^{TR}_{\lambda}= argmin J(\mathbf{w})
\end{equation}
\noindent We now introduce  the \emph{local } $\mathcal{CTR}$  functionals which  describes the local  problems on each sub-domain $\Omega_i$.

\begin{Definition}
Let $\Omega=\bigcup_{i=1}^{p} \Omega_i $ be the domain decomposition in (\ref{decom}).  For any vector $\mathbf{w}\in \Re^N$, let:
$J_{\Omega_i}= RO_i[J]$  be the operator as defined in (\ref{restrfun}) and let
$$ \tilde{J}_{\Omega_{ij}}: \mathbf{w}^{RO_{i}} \mapsto \tilde{J}_{\Omega_{ij}}(\mathbf{w}^{RO_{i}})\in \Re$$
be a  quadratic operator defined in $\Re^{card(\Omega_{ij})}$. The operator
\begin{equation}\label{min3dec}
J_{\Omega_i}(\mathbf{w}^{RO_i}):=  J_{\Omega_i}(\mathbf{w}^{RO_{i}}) +\omega_i \tilde{J}_{\Omega_{ij}}(\mathbf{w}^{RO_{ij}})
\end{equation}
is  the local $\mathcal{CTR}$ functional defined on $\Omega_i$ and on $\Omega_i\cap \Omega_j$. Parameters $\omega_i$ are  local regularization parameters. Then
 \begin{equation}\label{min3dec-prob}
 \mathbf{w}^{TR_i}_{\lambda,\omega_i}:= argmin_{\mathbf{w}^{RO_i}} J_{\Omega_i}(\mathbf{w}^{RO_i})\quad .
 \end{equation}
is the solution of the local TR problem. Since the local  functional is quadratic this solution is also unique, once index $i$ has been fixed.

\end{Definition}
\noindent \emph{Remark}: From (\ref{DD-J})  it follows that
\begin{equation}\label{DD-JL}
 J_{D^3L}:= \sum_{i=1,p}J^{EO_i}_{\Omega_i}= \underbrace{\sum_{i=1,p}J^{EO_i}_{\Omega_i}}_{J}+\underbrace{\sum_{i=1,p}\omega_i \tilde{J}_{\Omega_{ij}}^{EO_i}}_{\mathcal{C}}.
\end{equation}
In practice, $J_{D^3L}$ is obtained from a restriction of the $\mathcal{CTR}$ functional $J_{\Omega}$ in (\ref{min3glob}), and adding  a \emph{local} functional  defined on the overlapping regions in $\Omega_{ij}$. This is done in order  to enforce a sufficient continuity of local  solutions  onto the overlap region between  adjacent domains $\Omega_i$ and $\Omega_j$.
Operator $\tilde{J}_{\Omega_{ij}}$ can be suitably defined according to the specific requirements of the solution of the $\mathcal{CTR}$ problem. \\

\noindent The following  result relates  the solution of $\mathcal{CTR}$ problem in (\ref{min3glob}) to the solutions of the local TR problems in (\ref{min3dec}). From simplicity of notations, in the following Theorem we assume that $\mathcal{C}$ is  quadratic functional. The result still holds if this is a more general convex functional.
\begin{Theorem}
Let
$$\Omega= \bigcup_{i=1,p} \Omega_i$$ be a domain decomposition of $\Omega$ defined in (\ref{decom}), and let (\ref{DD-JL})
be  the associated functional decomposition.   Then let $\mathbf{w}^{TR}_{\lambda}$ be  defined in (\ref{min3glob})
and let $\widehat{\mathbf{w}}^{TR}_{\lambda}$ be defined as follows:
$$\widehat{\mathbf{w}}^{TR}_{\lambda}=\sum_i ({\mathbf{w}}_{\lambda,\omega_i}^{TR_i})^{EO_i}.$$
It is:
$$ \widehat{{\mathbf{w}}}^{TR}_{\lambda} = \mathbf{w}^{TR}_{\lambda}\,\,.$$
\begin{proof} $J_{ \Omega}$ is convex, as well as all the functionals ${J}_{\Omega_i}$, so their (unique) minimum,   $\mathbf{w}^{TR}_{\lambda}$ and ${\mathbf{w}}^{TR_i}_{\lambda,\omega_i}$, respectively, are obtained as zero of their  gradients, i.e.:
\begin{equation}\label{uno}
     \nabla J[ \mathbf{w}_{\lambda}^{TR}]= 0 \quad, \quad \nabla  {J}_{\Omega_i}[ {\mathbf{w}}^{TR_i}_{\lambda,\omega_i}]=0 \,\,.
\end{equation}
From (\ref{min3dec}) it follows that
\begin{equation}\label{due}
    \nabla {J}_{ \Omega_i}[{\mathbf{w}}^{TR_i}_{\lambda,\omega_i}] = \nabla J_{\Omega_i}[{\mathbf{w}}^{TR_i}_{\lambda,\omega_i}].
\end{equation}
From (\ref{exten}) it is:
\begin{equation}\label{tre}
     {\mathbf{w}}^{RO_i}_{\lambda,\omega_i}= \sum_{j=1,p}({\mathbf{w}}_{\lambda,\omega_i}^{TR_j})^{EO_j}, \quad on \quad \Omega_i\,\,.
\end{equation}
From (\ref{uno}), (\ref{due}) and  (\ref{tre}),   it follows that
\begin{equation}\label{quattro}
    0= \nabla J_{\Omega_i}({\mathbf{w}}_{\lambda,\omega_i}^{TR_i})=\nabla J_{\Omega_i}^{EO_i}\left (\sum_{j=1,p} ({\mathbf{w}}_{\lambda,\omega_i}^{TR_i})^{EO_i}\right ).
\end{equation}
\noindent By summing each equation in (\ref{quattro}) for $i=1, \ldots,p$ on all sub-domains $\Omega_i$,  from (\ref{quattro}) it follows that:
\begin{equation}\label{1}
    \sum_{i} \nabla J_{\Omega_i}^{EO_i} \left (\sum_j ({\mathbf{w}}_{\lambda,\omega_j}^{EO_j})^{TR}\right )=0 \Leftrightarrow \sum_{i} \nabla J_{\Omega_i}^{EO_i} (\hat{\mathbf{w}}^{TR}_{}\lambda)=0.
\end{equation}
From  the linearity of the gradients of $J_{\Omega_i}$, it is
\begin{equation}\label{2}
    \sum_{i} \nabla J_{\Omega_i}^{EO_i}(\widehat{\mathbf{w}}_{\lambda}^{TR})= \nabla \sum_{i} J_{\Omega_i}^{EO_i}(\widehat{\mathbf{w}}_{\lambda}^{TR}) =\nabla J(\widehat{\mathbf{w}}_{\lambda}^{TR})\,\,.
\end{equation}
 \noindent Hence,  from (\ref{2}) it follows
$$ \sum_i \nabla J_{\Omega_i}(\hat{\mathbf{w}}^{TR}_{\lambda})= 0 \Leftrightarrow \nabla J(\hat{\mathbf{w}}^{TR}_{\lambda}) = 0\,\,.$$
Finally,
$$\nabla J (\widehat{\mathbf{w}}_{\lambda}^{TR}) =0 \Rightarrow \widehat{\mathbf{w}}_{\lambda}^{TR} \equiv \mathbf{w}_{\lambda}^{TR}\,\,,$$
where the last equality holds because the minimum is unique.
\end{proof}
\end{Theorem}
\noindent   In conclusion, the minimum of $J_{\Omega}$,  can be  obtained by patching together the minima of  ${J}_{\Omega_i}$. This means that the accuracy per parameter metric is highly improved in this way.

\section{The D$^3$L parallel algorithm}

\begin{Definition}
Let $\mathcal{A}^{loc}_{D^3L}$ denote the algorithm solving the local $\mathcal{CTR}$ functional. Parallel  D$^3$L  algorithm  is symbolically denoted as 
\begin{equation}\label{D-TR:algorithm}
\mathcal{A}^{D^3L}:=\bigcup_{i=1,p} \mathcal{A}^{loc}_{D^3L}(\Omega_i).
\end{equation}
\begin{flushright}
$\spadesuit$
\end{flushright}
\end{Definition}
\noindent Parallel D$^3$L algorithm is described by \textbf{Algorithm 1}.  Similarly, the  $\mathcal{A}^{loc}_{D^3L}$ is described by \textbf{Algorithm 2}. $\mathcal{A}_{D^3L}^{loc}$ computes a local minimum of $\mathbf{J}_{\Omega_i}$ by solving the normal equations arising from  the linear least squares (LLS) problem. we note, in line 6, the exchange of $\mathbf{w}_{i}^l$. This is done in order  to enforce a sufficient continuity of local  solutions  onto the overlap region between  adjacent domains $\Omega_i$ and $\Omega_j$.\\

\begin{algorithm}{Algorithm 1 \label{DD-TRAlg}}
\begin{algorithmic}[1]
\Procedure{$D^3L$}{$in: \mathbf{A}, \mathbf{y}; out: \mathbf{w^{\mathbf{TR}}_{\lambda}}$}
\State {\% Domain Decomposition Step}
\Repeat
\State \hspace*{.1cm} $l:=l+1$
\State \hspace*{.2cm}{\bf Call } Loc\_LLS $(in: \mathbf{A}^{RO_i}, \mathbf{y}^{RO_i}; out: \mathbf{w}^l_{i})$
\State \hspace*{.2cm}\textbf{Exchange} $\mathbf{w}_{i}^l$ between adjacent subdomains
\Until{$\|\mathbf{w}_{i}^l - \mathbf{w}_{i}^{l-1}\|< eps$}

\State {\% End  Domain Decomposition Step}
\State {\bf Gather} of $\mathbf{w}_{i}^l: \mathbf{w}^{\mathbf{TR}}= 
{\arg\min}_{i} \left\{J \left (\mathbf{w}_{i}^l\right ) \right \}$
 \label{puntoparall2}
 \EndProcedure
\end{algorithmic}
\end{algorithm}

\normalsize

\begin{algorithm}{Algorithm 2 }\label{local-TRAlg:LLS}
\begin{algorithmic}[1]
\Procedure{Loc-LLS}{$\mathbf{A^{RO_{i}}}, \mathbf{y}^{RO_i}; out: \mathbf{u}^l_{i}$}
\State {\bf Initialize} $l:=0$;
\State \textbf{repeat}
\State \hspace*{.5cm}{\bf Compute} $\delta \mathbf{{u}}_{i}^l=\arg\min \,\mathbf{J}_{\Omega_{i}}$ by solving the normal equations system
\State \hspace*{.5cm}{\bf Update} $\mathbf{{u}}_{i}^l=\mathbf{{u}}_{i}^l+\delta \mathbf{{u}}_{i}^l $
\State \hspace*{.5cm}{\bf Update} $l=l+1$
\State \textbf{until} (convergence is reached)
 \EndProcedure
\end{algorithmic}
\end{algorithm}

\subsection{Performance Analysis}
\noindent  We  use time complexity and scalability as performance metrics. Our aim is to highlight the benefits arising from using the  decomposition approach instead of solving the problem on the whole domain. 
\noindent  \begin{Definition}
A uniform  decomposition of $ \Omega$ is such that  if we let $$size(\Omega)=  N $$ be the size of the whole data domain, then each subdomain $\Omega_i$ is such that  $$size( \Omega_i)=  N_{loc}, \quad i=1,\ldots,p ,$$
where $N_{loc}=\frac{N}{p}\geq 1$.
\begin{flushright}
$\spadesuit$
\end{flushright} 
 \end{Definition}
\noindent  Various metrics have been developed to
assist in evaluating the scalability  of a parallel algorithm:  speedup, 
throughput,  efficiency are the most used. Each one highlights  specific needs and limits to be answered by the parallel software. Since we  mainly focus on the algorithm's scalability  arising from the  proposed framework,  we  consider the so-called  scale-up factor first introduced in  \cite{DD-DA}.\\
\noindent  Let  $T(\mathcal{A}^{D^3L}(\Omega))$ denote  time complexity of  $\mathcal{A}^{D^3L}( \Omega)$.   The major computational task to be performed  are as follows:
\begin{enumerate}
  \item Computation of the kernel  $RO_{ji}[\mathbf{A}]$ (the time complexity of such an operation scales as $LN^2$ ) 
  \item Solution of the normal equations, involving at each iteration two matrix-vector products with $RO_{ji}[(\mathbf{A}^T)]$   and $RO_{ji}[\mathbf{A}]$ (whose time complexity scales as $L^2N^4$).
\end{enumerate} 

\noindent We pose $d=4$. We now provide an estimate of the time complexity of each local algorithm, denoted as $T(\mathcal{A}^{Loc}(\Omega_i))$. 
\noindent The first result straightforwardly derives from the definition  of the scale-up factor: 
\noindent \begin{Proposition}[ Scale-up factor]
The (relative) scale-up factor of  $\mathcal{A}^{D^3L}_{}(\Omega)$ related to $\mathcal{A}^{loc}_{}(\Omega_i)$, denoted as
$Sc_{p}(\mathcal{A}^{D^3L}_{}(\Omega))$, is  $$Sc_p(\mathcal{A}^{D^3L}( \Omega)):=\frac{1}{p}\times\frac{T(\mathcal{A}^{D^3L}_{}( \Omega))}{ T(\mathcal{A}^{loc}_{}( \Omega_i))}\,\,\,, $$
where $p$ is the number of subdomains. It is

\begin{equation}\label{scaleup}
  Sc_{p}(\mathcal{A}^{D^3L})\geq  \alpha(N_{loc},p)\,(p)^{d-1},
\end{equation}
where
$$\alpha(N_{loc},p)=\frac{a_d+a_{d-1}\frac{1}{N}+\ldots+\frac{a_0}{N_{loc}^{d}}}{a_d+a_{d-1}\frac{p}{N_{loc}}+\ldots+\frac{a_0(p)^{d}}{N_{loc}^{d}}}  \quad $$
and
$$ \lim_{p\rightarrow N_{loc}} \alpha(N_{loc},p)=\beta \in]0,1].$$
\begin{flushright}
$\spadesuit$
\end{flushright}
\end{Proposition}

\begin{Corollary}
If $ a_i= 0 \quad \forall i\in[0,d-1]$, then $\beta=1$, that is, 
$$ \lim_{p\rightarrow N_{loc}} \alpha(N_{loc},p)=1.$$
Then,
$$\lim_{N_{loc}\rightarrow \infty} \alpha(N_{loc},p)=1.$$
\begin{flushright}
$\clubsuit$
\end{flushright}

\end{Corollary}

\begin{Corollary}\label{corscaleup}
If $N_{loc}$ is fixed, then
$$ \lim_{p\rightarrow N_{loc} }Sc_{p}(\mathcal{A}^{D^3L})= \beta\cdot N_{loc}^{d-1} \quad ;$$
while if $p$ is fixed, then
$$ \lim_{N_{loc}\rightarrow \infty }Sc_{p}(\mathcal{A}^{D^3L})= const \neq 0\quad . $$
\begin{flushright}
$\clubsuit$
\end{flushright}

\end{Corollary}

\noindent  From (\ref{scaleup}) it results that, considering one iteration of the whole parallel algorithm,  the growth of the scale-up factor essentially is one order less than the time complexity of the reduced kernel. In other words, the time complexity of the reduced kernel impacts mostly the scalability of the parallel algorithm.  In particular, since parameter $d$ is equal to $4$, it follows that the asymptotic scaling factor of the parallel algorithm, with respect to $p$, is bounded above by three.

\noindent Besides the time complexity, scalability is also affected by the  communication overhead of the parallel algorithm. The surface-to-volume ratio is a measure of the amount of data exchange (proportional to surface area of domain) per unit operation (proportional to volume of domain). It is straightforward to prove that the surface-to-volume ratio of the uniform  decomposition of  $ \Omega$ is
\begin{equation}\label{surfacetovolume}
 \frac{\mathcal{S}}{\mathcal{V}}= \mathcal{O}\left (\frac{1}{N_{loc}}\right) \quad .
\end{equation}

\noindent 
Theoretical performance analysis in terms of the scale up factor tell us that  if the application  needs to reach a prescribed time-to-solution (strong scaling), we can exploit the high performance of emerging GPU. In this case,  $p$ can increase while $N_{loc}$ is fixed according to memory constraints of GPU;  the scale-up factor increases with a fixed surface-to-volume ratio. On the other hand, for computationally intensive applications, it is preferable to exploit the weak scaling of  clusters of distributed memory multiprocessors. In fact, by fixing $p$  while $N_{loc}$ may increase according to application requirements, the scaling factor is kept constant, while the surface-to-volume ratio decreases supporting overall efficiency. Summarizing,  performance analysis tells us that looking at the scale up factor one may find the appropriate mapping of the specific application on the target  architecture.

\section{VALIDATION}

\noindent  The proposed  approach  has a straightforward application to Convolution Neural Networks (CNN) which are specialized in processing data that has a grid-like topology, such as an image. In this case, the domain $\Omega$ is the image while the core building block of the  CNN is the convolutional layer. This layer performs a dot product between two matrices, where one matrix is the kernel and the other matrix is a restricted portion of the image. As a result, by decomposing $\Omega$ according to the standard two-dimensional block cyclic distribution  of the image matrix, which is implemented in linear algebra libraries for parallel matrix computation (MATLAB, ScaLAPACK, PBLAS, ...), we    get to the standard  data layout of dense matrix on distributed-memory architectures which permits the use of the Level 3 of BLAS during computations on a single node. \\
We validate the algorithm on the DIGITS dataset of MATLABR2014b, by using the high-performance hybrid computing architecture of the SCoPE (Sistema Cooperativo Per Elaborazioni scientifiche multidiscipliari) data center, located at University of Naples Federico II. The architecture is composed of 8 nodes consisting of distributed memory DELL M600 blades. The blades are connected by 10 gigabit Ethernet technology and each of them is composed of 2 Intel Xeon@2.33GHz quadcore processors sharing the same local 16 GB RAM memory for a number of 8 cores per blade and of 64 total cores. We validate the algorithm using the Parallel Computing Toolbox of MATLABR2014b.  In Table \ref{my-labelFinal} we report strong scaling results computed using the scale-up factor $Sc_{p}(\mathcal{A}^{D^3L}_{}(\Omega))$ as given in Proposition 21 and, just to make a comparison with a standard metric, we also report the classical speed-up metric indicated as $S_p(\mathcal{A}^{D^3L}_{}(\Omega))$. Weak scaling scalability is reported in Table 2 using the scale-up factor.




\normalsize

\begin{table}
\centering
\begin{tabular}{cccccccc}
\hline 
  $p$  & $ N$    &  & & $T(\mathcal{A}^{D^3L}_{}( \Omega))$ & &  \\
  
   \hline \\
   
  1  & $ 1.23 \times 10^4$&  & &$1.56\times 10^3 $&  &   \\

\hline \hline
   & $\frac{S}{V}$ & $T(\mathcal{A}^{loc}_{}( \Omega_i))$ & $T^{p}(\mathcal{A}^{D^3L}_{}( \Omega))$ & $S_p(\mathcal{A}^{D^3L}_{}(\Omega))$& $Sc_{p}(\mathcal{A}^{D^3L}_{}(\Omega))$  \\
 
\hline\\

  2 &   $1.06 \times 10^{-7} $  &$9.8\times 10^1$ &$0.8 \times 10^3$ &$1.8$& $1.5$ \\

    4&     $6.7 \times 10^{-5} $ &$3.0\times 10^1$ &$0.48\times 10^3 $&$3.2$ &$6.5$  \\
   
8&     $2.5 \times 10^{-5} $  &$2.7\times 10^{-1}$ &$0.24 \times 10^3 $&$6.5$ &$7.2$\\
16&     $1.4 \times 10^{-5} $ &$1.2\times 10^{-2}$ &$0.1\times 10^3 $&$14.2$  &$8.1$\\
\hline 
\end{tabular}
\caption{Strong scaling results  }\label{my-labelFinal}
\end{table}

\begin{table}
\centering
\begin{tabular}{c|cccccc}
\hline
         $p$  & 2     & 4 & 8   & 16 &32 &64  \\
         
          \hline\\											

        $N$  & $3.07\times 10^3$     & $1.23\times 10^4$ & $4.90\times 10^4$   & $1.97\times 10^5$ &$7.86\times 10^5$ &$3.15\times 10^6$  \\

         \hline\\

$Sc_{p}(\mathcal{A}^{D^3L})(\Omega)$ &   $ 5.96 \times 10^0$    & $2.39\times 10^1$ & $9.54\times 10^1$   & $3.81\times 10^2$ &  $1.53\times 10^3$ & $6.10\times 10^3$ \\
\hline
\end{tabular}
\caption{Weak Scaling scalability  of  $\mathcal{A}^{D^3L}(\Omega)$. }\label{my-label2}
\end{table}

\section{Future works}

\noindent We underline that  we refer to DKL problems which give rise to convex optimization. More in general,  we can address DL problems where, in case of non convex loss function, we  consider a surrogate convex  loss function.  An example of such surrogate loss functions is the hinge loss, $\Phi(t)=max(1- t,0)$, which is the loss used by Support Vector Machines SVMs. Another example is the logistic loss, $\Phi(t)=1/(1+exp(-t))$, used by the logistic regression model.  A natural questions to ask is how much have we lost by this change. The property of whether minimizing the surrogate function leads to a function that also minimizes the loss function is often referred to as consistency \cite{Bartlett}. This property will depend on the surrogate function. One of the most useful characterizations was given in \cite{Bartlett} and states that if $\Phi$ is convex then it is consistent if and only if it is differentiable at zero and $\Phi'(0)<0$. This includes most of the commonly used surrogate loss functions, including hinge, logistic regression and Huber loss functions. \\
\noindent Following \cite{DamoreCacciapuoti,DamoreCostantinescu} we suggest the employment of the proposed approach on the plentiful  literature of  Physical Informed Neural Network (PINN) applications, where data are constrained according to a specific  physical model generally modelled by evolutive Partial Differential Equations.


\end{document}